\relax
\documentclass[letterpaper]{article} 
\usepackage{aaai22}  
\usepackage{times}  
\usepackage{helvet}  
\usepackage{courier}  
\usepackage[hyphens]{url}  
\usepackage{graphicx} 
\usepackage{natbib}  
\usepackage{caption} 
\DeclareCaptionStyle{ruled}{labelfont=normalfont,labelsep=colon,strut=off} 
\usepackage{algorithm}
\usepackage{algorithmic}
\usepackage{xcolor}
\usepackage{tcolorbox}
\usepackage{adjustbox}
\usepackage{subfigure}
\usepackage{booktabs}
\usepackage{multirow}

\input{Definitions}

\usepackage{newfloat}
\usepackage{listings}
\floatstyle{ruled}
\newfloat{listing}{tb}{lst}{}
\floatname{listing}{Listing}
\title{A Generic Approach for Enhancing GANs by Regularized Latent Optimization}
\author{
    Yufan Zhou\textsuperscript{\rm 1}, Chunyuan Li\textsuperscript{\rm 2}, Changyou Chen\textsuperscript{\rm 1}, Jinhui Xu\textsuperscript{\rm 1}\\
}
\affiliations{
    \textsuperscript{\rm 1}State University of New York at Buffalo, ~~~\textsuperscript{\rm 2} Microsoft Research, Redmond\\

    \{yufanzho, changyou, jinhui\}@buffalo.edu,~~~ chunyl@microsoft.com    
}

\newcommand{\RN}[1]{%
	\textup{\lowercase\expandafter{\it \romannumeral#1}}%
}

\DeclareMathOperator*{\argmin}{arg\,min}

\def\Div{\textsf{Div}}
\usepackage{bibentry}

\begin{document}

\maketitle

\begin{abstract}
With the rapidly growing model complexity and data volume, training deep generative models (DGMs) for better performance has becoming an increasingly more important challenge. Previous research on this problem has mainly focused on improving DGMs by either introducing new objective functions or designing more expressive model architectures. However, such approaches often introduce significantly more computational and/or designing overhead. To resolve such issues, we introduce in this paper a generic framework called {\em generative-model inference} that is capable of enhancing pre-trained GANs effectively and seamlessly in a variety of application scenarios. Our basic idea is to efficiently infer the optimal latent distribution for the given requirements using Wasserstein gradient flow techniques, instead of re-training or fine-tuning pre-trained model parameters. Extensive experimental results on applications like image generation, image translation, text-to-image generation, image inpainting, and text-guided image editing suggest the effectiveness and superiority of our proposed framework. 
\end{abstract}

\section{Introduction}

Deep generative models (DGMs) are playing an important role in deep learning, and have been used in a wide range of  applications, such as image generation \cite{goodfellow2014generative, brock2018large, karras2019style, miyato2018spectral}, image translation \cite{ zhu2017unpaired, huang2018multimodal, choi2020stargan}, text-to-image generation \cite{nguyen2017plug, xu2018attngan, qiao2019mirrorgan, ramesh2021zero}, etc. An 
emerging challenge in this topic is how to design scalable DGMs that can accommodate industrial-scale and high-dimensional data.
Most of previous research on this problem has
focused on designing either better model objective functions or more expressive model architectures. For example, the recent denoising diffusion probabilistic models (DDPMs) \citep{NEURIPS2020_4c5bcfec} and its variants \citep{song2021denoising, song2021scorebased} employ carefully designed objectives to 
enable the model to progressively generate images from noises (by simulating the denoising process).  
The work in \citep{gong2019autogan,Gao_2020_CVPR} adopts 
the idea of neural architecture search, 
instead of using some pre-defined architectures,
to obtain
significant improvement over baseline models. Despite
achieving impressive results, such models also share a common 
issue, that is, they may incur  
significantly more computational and/or designing overhead. For instance, DDPM is trained on TPU v3-8 that has 128 GB memory, while a typical generative adversarial network (GAN) can be trained only on one GPU with less than 12 GB memory. To generate 50,000 images of $32\times 32$ resolution on a single Nvidia RTX 2080 Ti GPU, DDPM needs around 20 hours while GAN needs only one minute \citep{song2021denoising}.

To resolve this issue, we explore a new direction and propose a generic framework called {\em generative-model inference} that allows us to directly enhance existing pre-trained generative adversarial networks (GANs) ({\it i.e.}, without changing their pre-trained model parameters), and only need to optimize the latent codes. 
This is somewhat similar to
Bayesian inference where one aims to infer the optimal latent code of an observation in a latent variable model, and hence motivates us to call 
our framework {\em generative-model inference}. Our basic idea is to formulate the problem as a Wasserstein gradient flow (WGF) \citep{ambrosio2008gradient, santambrogio2016}, which optimizes the generated distribution (through the latent distribution) in the space of probability measures. To solve the WGF problem, a regularized particle-based solution and kernel ridge regression are developed. Our framework is quite general and can be applied to different application scenarios. For example, we have conducted experiments in applications including image generation, image translation, text-to-image generation, image inpainting, and text-guided image editing, and obtain state-of-the-art (SOTA) results. 
Our main contributions can be summarized as follows.
\begin{itemize}
    \item We propose a generic framework called generative-model inference to boost pre-trained generative models. Our framework is generic and efficient, which includes some existing methods as special cases and can be applied to applications not considered by related works.
    \item We propose an efficient solution for our framework by leveraging new density estimation techniques and the WGF method. 
    \item We conduct extensive experiments to illustrate the wide applicability and effectiveness of our framework, improving the performance of the SOTA models.
\end{itemize}

\section{The Proposed Method}



\begin{figure}[t!]
    \begin{minipage}{\linewidth}
      \begin{algorithm}[H]
        \caption{Generative-Model Inference}\label{algo:sampling}
            \begin{algorithmic}[1]
                    \REQUIRE A pre-trained generator $g$ and a feature extractor $d$, mollifier distribution $\psi(\epsilonb)$, step size $\lambda_1, \lambda_2, \lambda_3$, condition set $c$.
                    \STATE Sample latent codes $\{\zb^0_i\}_{i=1}^n$, generate samples $\{\xb^0 _i=g(\zb^0_i)\}_{i=1}^n$
                    \IF{Having access to training dataset}
                        \STATE Sample real samples $\{\xb^\prime _i\}_{i=1}^n$;
                    \ENDIF
                    \FOR{$t=0, 1, ..., T-1$}
                        \STATE {\color{red} // Gradient estimation}
                        \STATE Estimate $A^t_i = \log q_{\thetab}(\xb^t_i\vert c)$ via \eqref{eq:grad_log_q}; 
                        \STATE If having access to training dataset, estimate $B^t_i = \log p(\xb^t_i)$ via \eqref{eq:grad_log_p}; Else $B^t_i  = \mathbf{0}$.
                        \STATE If having access to $p(c \vert \xb)$, estimate $C^t_i = \log p(c \vert \xb^t_i)$; Else  $C^t_i = \mathbf{0}$.
                        \STATE {\color{red} // Generative-model inference to update latent code}
                        \STATE $\zb^{t+1}_i \leftarrow \zb^{t}_i - \lambda_1 \nabla_{\zb^t_i} A^t_i + \lambda_2 \nabla_{\zb^t_i} B^t_i +  \lambda_3\nabla_{\zb^t_i} C^t_i$
                        \STATE Generate samples $\{\xb^{t+1} _i\}_{i=1}^n$ where $\xb^{t+1}_i = g(\zb^{t+1}_i)$.
                    \ENDFOR
                    \RETURN $\{\xb^T _i\}_{i=1}^n$.
            \end{algorithmic}
        \end{algorithm}
    \end{minipage}
\end{figure}
  
\subsection{Problem Setup}
We use $\xb \in \mathcal{X}$ and $\zb \in \mathcal{Z}$ to denote a data sample and its corresponding latent representation/code  in the data space $\mathcal{X}$ and latent space $\mathcal{Z}$, respectively. Assume that a pre-trained parameterized mapping $g_{\thetab}: \mathcal{Z} \rightarrow \mathcal{X}$ is induced by a neural network with parameter $\thetab$, which takes a latent code $\zb$ as input and generates a data sample $\xb$. We use $p(\xb)$, $q_{\thetab}(\xb)$ to denote the distribution of the training and generated samples, respectively. Typically, one has only samples from  $p(\xb)$ and $q_{\thetab}(\xb)$ but not their distribution forms. 

A standard deep generative model is typically defined by the following generation process: $\zb \sim q^0(\zb), \xb \sim q_{\thetab}(\xb|\zb)$, where $q^0$ denotes a simple prior distribution such as the standard Gaussian distribution, and $q_{\thetab}(\xb|\zb)$ denotes the conditional distribution induced by the generator network\footnote{In some cases such as GANs, the mapping from $\zb$ to $\xb$ can be deterministic, and thus the corresponding conditional distribution $q_{\thetab}(\xb|\zb)$ will be a Dirac delta function.}. Our proposed generative-model inference framework is defined on a more general setting by considering extra conditional information, denoted as $c$ (to be specified below). In other words, instead of modeling the pure generated data distribution $q_{\thetab}(\xb)$, we consider $q_{\thetab}(\xb|c)$. It is worth noting that when $c$ is an empty set, it recovers the standard unconditional setting. To explain our proposed generative-model inference problem, we use Bayes theory to rewrite the generative distribution as: 
\begin{align}\label{eq:gminference}
    q_{\thetab}(\xb|c) = \int_{\mathcal{Z}}q_{\thetab}(\zb|c)q_{\thetab}(\xb|\zb)\mathrm{d}\zb~,
\end{align}
where $q_{\thetab}(\zb|c)$ is the conditional distribution of the latent code $\zb$ given $c$. $q_{\thetab}(\zb \vert c)$ enables the latent codes to encode the desired information to better guide new sample generation.
In our experiments, we consider the following cases of $c$: 1) {\bf Standard image generation}: $c$ corresponds to the information of training dataset; 2) {\bf Text-to-image generation}: $c$ corresponds to the text information describing desired attributes; 3) {\bf Image translation}: $c$ contains information of the target domain, which could be the label of the target domain; 4) {\bf Image inpainting}: $c$ represents the information of unmasked region of the masked image; 5) {\bf Text-guided image editing}: $c$ contains both text description and the original image. 
It is worth noting that the aforementioned related works \citep{ansari2020refining, che2020your, tanaka2019discriminator, zhou2021learning} do not consider the conditional distribution extension with $c$, thus can not deal with related tasks.

With our new formulation \eqref{eq:gminference}, a tempting way
to generate a sample is to use the following procedure   
$\zb \sim q_{\thetab}(\zb|c), \xb \sim q_{\thetab}(\xb|\zb)$,
instead of adopting the standard process. 
This intuitively means that we first find the desired latent distribution from $q_{\thetab}(\zb|c)$ given the conditions and requirements, and then generate the corresponding samples from the conditional generator $q_{\thetab}(\xb|\zb)$. Unfortunately, 
such a generating procedure can not be directly applied in practice as $q_{\thetab}(\zb|c)$ is typically unknown. The main goal of this paper is thus to design effective algorithms to infer $q_{\thetab}(\zb|c)$, termed as generative-model inference. 

\begin{definition}
{\bf Generative-Model Inference} is defined as the problem of inferring $q_{\thetab}(\zb|c)$ in \eqref{eq:gminference}.
\end{definition}
\begin{remark}[Why generative-model inference]
There are at least two reasons: 1) DGMs such as GANs are well known to suffer from the mode-collapse problem, partially due to the adoption of a simple prior distribution for the latent code. With a simple prior, it might be hard for the conditional distribution induced by GAN to perfectly fit a complex data distribution. This is evidenced in \cite{zhou2021learning}. Our framework, by contrast, can adaptively infer an expressive latent distribution for the generator according to the given condition $c$, which can consequently lay down the burden of directly modeling a complex conditional distribution. 2) With generative-model inference, one no longer needs to fine-tune or re-train the parameters of GAN to obtain better results. The proposed method, by contrast, can be directly plugged into pre-trained SOTA models, leading to an 
efficient way.
\end{remark}

Together with $q_{\thetab}(\xb|\zb)$ from a pre-trained generative model, we first give 
an overview on how to use our generative-model inference to generate better samples ({\it i.e.}, Algorithm \ref{algo:sampling}). Detailed derivation of the algorithm is given in the following sections.

\subsection{Generative-Model Inference  with Wasserstein Gradient Flows}
With the extra information $c$, 
we consider to model the unknown ground-truth distribution $p(\xb \vert c)$, instead of the previously defined $p(\xb)$ in standard DGMs. Our goal is to make the generated distribution match $p(\xb\vert c)$.
We propose to solve the problem by minimizing the KL-divergence $\mathcal{D}_{KL}(q_{\thetab}(\xb\vert c), p(\xb \vert c))$ w.r.t $q_{\thetab}(\zb\vert c)$ in the space of probability measures. The problem can be formulated as a Wasserstein gradient flow (WGF), which ensures maximal decrease of the KL-divergence when $q_{\thetab}(\xb\vert c)$ evolves towards the target distribution $p(\xb \vert c)$ in the space of probability measures. Theorem~\ref{lemma:wgf_kl} stands as the key result of formulating $q_{\thetab}(\xb\vert c)$ as a WGF. All proofs are provided in the Appendix. Note that although we only consider minimizing the KL-divergence, our framework can be naturally extended to any $f$-divergence, as discussed in the Appendix.

\begin{theorem}\label{lemma:wgf_kl}
Let $\Div$ denote the divergence operation \citep{santambrogio2016}, and $p(c \vert \xb)$ denote the likelihood of $c$ given $\xb$. The Wasserstein gradient flow of the functional $\mathcal{F}(q) = \mathcal{D}_{KL}(q_{\thetab}(\xb\vert c), p(\xb \vert c))$ can be represented by the  partial differential equation (PDE):
\begin{align}
    \partial q_{\thetab}(\xb\vert c)/&\partial t =  \Div (q_{\thetab}(\xb\vert c) \nabla_{\xb} \log q_{\thetab}(\xb\vert c) \nonumber \\& - q_{\thetab}(\xb\vert c)\nabla_{\xb} \log p(\xb) - q_{\thetab}(\xb\vert c) \nabla_{\xb} \log p(c \vert \xb)) \nonumber
\end{align}
Furthermore, the associated ordinary differential equation (ODE) for $\xb$ is:
\begin{align}\label{eq:wgf_sde}
    \text{d}\xb =  \nabla_{\xb} \log p(\xb) \text{d}t + \nabla_{\xb} \log p(c \vert \xb) \text{d}t -\nabla_{\xb} \log q_{\thetab}(\xb\vert c) \text{d}t
\end{align}
\end{theorem}

Theorem~\ref{lemma:wgf_kl} provides the evolving direction of the generated distribution $q_{\thetab}(\xb \vert c)$, which consequently translates to\footnote{$q_{\thetab}(\zb|c)$ relates to $q_{\thetab}(\xb \vert c)$ through \eqref{eq:gminference}. As illustrated in Algorithm~\ref{algo:sampling}, we use particles to approximate $q_{\thetab}(\zb|c)$; thus solving $q_{\thetab}(\zb|c)$ corresponds to updating the particles of the latent codes.} $q_{\thetab}(\zb|c)$. The updates are performed by approximating the distributions with particles and applying gradient-based methods ({\it i.e.}, via time discrtization of \eqref{eq:wgf_sde} to update the particles). However, the main challenge of such an approach comes from the fact that $q_{\thetab}(\xb \vert c), p(\xb), p(c\vert \xb)$ are all unknown, thus hindering direct calculations for $\nabla_{\xb}\log q_{\thetab}(\xb \vert c)$, $\nabla_{\xb}\log p(\xb)$, and $\nabla_{\xb}\log p(c\vert \xb)$ in \eqref{eq:wgf_sde}. In the following, we propose effective approximation techniques to estimate these quantities. 

\subsection{Gradient Estimations}
\paragraph{Estimating $\nabla_{\xb}\log q_{\thetab}(\xb|c)$:}
Our idea is to formulate $q_{\thetab}(\xb|c)$ as a tractable expectation form so that it can be approximated by samples. To this end, we assume that there exists an unknown function $r(\xb)$ such that $q_{\thetab}(\xb|c)$ can be decomposed as: $q_{\thetab}(\xb \vert c) = \int_{\mathcal{X}} r(\xb^\prime) k(\xb, \xb^\prime )\text{d}\xb^\prime$, where $k(\xb, \xb^\prime)$ is a kernel function, which we rewrite it as $k_{\xb}(\xb^\prime)$ for conciseness. In our implimentation, we construct the kernel as $k(\xb, \xb^\prime) = \exp(-\Vert d(\xb) - d(\xb^\prime)\Vert^2)$, where $d(\cdot)$ is some feature extractor, {\it e.g.}, the discriminator of GAN. For the purpose of computation stability (a brief discussion on why the regularized kernel is beneficial is provided in Appendix~\ref{app:regularizedkernel}), we further regularize the kernel with a pre-defined mollifier function $\psi$ \cite{craig2016blob} as: $k_{\psi}(\xb, \xb^\prime) = (k_{\xb}*\psi)(\xb^\prime)$, where ``$*$'' denotes the convolution operator. To enable efficient computation of the convolution, we restrict the mollifier function to satisfy the condition of $\int_{\mathcal{X}} \psi(\xb) \text{d} \xb=1$, {\it i.e.}, $\psi(\xb)$ is a probability density function (PDF) that  can be easily sampled from. Consequently, we can rewrite $q_{\thetab}(\xb \vert c)$ by replacing the kernel $k_{\xb}(\cdot)$ with the following approximate regularized kernel:
\begin{align}\label{eq:kernel}
    k_{\psi}(\xb, \xb^\prime) = (k_{\xb}*\psi)(\xb^\prime) = \int_{\mathcal{X}} k_{\xb}(\xb^\prime - \epsilonb)\psi(\epsilonb) \text{d}\epsilonb 
\end{align}
In practice, we use samples to approximate the convolution in \eqref{eq:kernel}, which leads to $\sum_{i=1}^m k(\xb, \xb^\prime - \epsilonb_i)/m, \text{ where } \epsilonb_i \sim \psi(\epsilonb)$. Below, we show that the error induced by the smoothing can actually be controlled.
\begin{theorem}\label{thm:bounded_error}
Assume that $k_{\xb}$ is twice differential with bounded derivatives, $\psi(\epsilonb)$ denotes a distribution with zero mean and bounded variance $\sigma^2 < \infty$. Let $k_{\psi}(\xb, \xb^\prime) = (k_{\xb} * \psi)(\xb^\prime)$ be the regularized kernel, then
$
\vert (k_{\xb} * \psi)(\xb^\prime) - k_{\xb}(\xb^\prime) \vert \leq  O(\sigma^2).
$
\end{theorem}
Next, we aim to re-write $q_{\thetab}(\xb|c)$ in an expectation form. To this end, we introduce $\alpha(\xb^\prime) = r(\xb^\prime)/q_{\thetab}(\xb^\prime\vert c)$ and reformulate $q_{\thetab}(\xb|c)$ as
\[
q_{\thetab}(\xb \vert c) = \mathbb{E}_{q_{\thetab}(\xb^\prime \vert c)}\left[\alpha(\xb^\prime) k_{\psi}(\xb, \xb^\prime )\right] \approx \sum_{i=1}^n \alphab_i k_{\psi}(\xb, \xb_i), 
\] 
where we use uniformly sampled $\{\xb_i\}_{i=1}^n$ from $q_{\thetab}(\xb \vert c)$ for approximation with $\alphab_i \triangleq \alpha(\xb_i)$. Let $\tilde{q}_{\thetab}(\xb \vert c)\triangleq \sum_{i=1}^n \alphab_i k_{\psi}(\xb, \xb_i)$, since $\alphab_i$'s are unknown, we propose to estimate them by enforcing $\tilde{q}_{\thetab}(\xb \vert c)$ to fit the ground-truth $q_{\thetab}(\xb \vert c)$. Denote $\alphab^\star$ as the collection of  optimal $\alphab_i$'s to approximate $q_{\thetab}(\xb \vert c)$. We can obtain $\alphab^\star$ by the following objective:
\[
\alphab^\star = \argmin _{\alphab \in R^n} \sum_{i=1}^n\Vert \tilde{q}_{\thetab}(\xb_i \vert c) - q_{\thetab}(\xb_i \vert c) \Vert^2/n.
\]
We further add an $L_2$-norm regularizer $\eta  \Vert \tilde{q} \Vert^2_{\mathcal{H}}$ with $\eta > 0$, where $\Vert \cdot \Vert_{\mathcal{H}}$ is the norm in the Reproducing Kernel Hilbert Space (RKHS) $\mathcal{H}$ induced by the kernel. As a result, we can obtain a closed-form solution for $\alphab^\star$ with kernel ridge regression (KRR): $\alphab^\star = (K^q + \eta I_n)^{-1}$, 
where $K^q \in R^{n \times n}$ denotes the kernel matrix with elements $K^q_{ij} = k_{\psi}(\xb_i, \xb_j)$, $I_n \in R^{n\times n}$ is the identity matrix.
By substituting $\alphab^\star$ in $\tilde{q}_{\thetab}(\xb \vert c)$ and following some simple derivations, we have the following approximation for $\log q_{\thetab}(\xb \vert c)$:
\begin{align}\label{eq:grad_log_q}
    \log q_{\thetab}(\xb \vert c) \approx \log \sum_{i=1}^n \sum_{j=1}^n H^q_{ji}k_{\psi}(\xb, \xb_i),
\end{align}
where $ H^q = (K^q + \eta I_n)^{-1}$. We can then use back-propagation to compute $\nabla_{\xb} \log q_{\thetab}(\xb \vert c) $ and $\nabla_{\zb} \log q_{\thetab}(\xb \vert c)$ where $\xb = g(\zb)$.
In the above approximation, if replacing $H^q$ with $\eta^{-1} I_{n}$, we essentially recover the kernel density estimator (KDE): $\log  q_{\thetab}(\xb \vert c) \approx \log \sum_{i=1}^n k_{\psi}(\xb, \xb_i)$. Thus, our method can be considered as an improvement over KDE by introducing pre-conditioning weights. The relation is formally shown in the following theorem.
\begin{theorem}\label{thm:krr_kde}
    Let $K$ be a $n \times n$ kernel matrix, $\eta>0$. Then the following holds: 
    \[
    \Vert (\eta I_{n} + K)^{-1} - \eta^{-1} I_{n}\Vert_{\mathcal{F}} \leq O(n\eta^{-4}),
    \]
    where $\Vert \cdot \Vert_{\mathcal{F}}$ denotes the Frobenius norm of matrix. In other words, when increasing the regularization level $\eta$, our proposed method approaches the kernel density estimator.
\end{theorem}

\paragraph{Estimating $\nabla_{\xb}\log p(\xb)$:}
Following a similar argument, $\log p(\xb)$ can be approximated, with samples $\{\xb_i^\prime\}_{i=1}^n$ from $p(\xb)$, as:
\begin{align}\label{eq:grad_log_p}
    \log p(\xb) \approx \log \sum_{i=1}^n \sum_{j=1}^n H^p_{ji}k_{\psi}(\xb, \xb_i^\prime),
\end{align}
where $ H^p = (K^p + \xi I_n)^{-1}$.

\paragraph{Estimating $\nabla_{\xb} \log p(c \vert \xb)$:}
The gradient will be problem dependent. 1) For image generation, $p(c|\xb)$ represents the likelihood of a generated sample $\xb$ belongs to the training dataset, whose information is encoded in $c$. Since the discriminator is trained to distinguish the real and fake images, we can approximate $p(c|\xb)$ with the pre-trained discriminator $d$, {\it i.e.}, $p(c \vert \xb) \propto e^{d(\xb)}$. Consequently, we have $\nabla_{\zb} \log p(c \vert g(\zb)) \approx \nabla_{\zb} d(g(\zb))$. This coincides with the update gradient proposed in \citep{che2020your}. Thus, \citep{che2020your}  can be viewed as a special case of our proposed method. The same result also applies to image translation, image completion and text-to-image generation tasks: In \cite{choi2020stargan,zhao2021comodgan,tao2021dfgan}, the discriminators are trained to classify whether the generated images satisfy the desired requirements, and thus can be directly used to estimate $p(c \vert \xb)$. 2) For text-related tasks using the pre-trained CLIP model \cite{radford2021learning} ({\it e.g.}, in one of our text-to-image generation model and text-guided image editing task), instead of using a pre-trained discriminator, we propose to approximate $p(c|\xb)$ directly from the output of CLIP. This is because the output of CLIP measures the semantic similarity between a given image-text pair and thus can be considered as a good approximation for the log-likelihood of $c$ given $\xb$.

\subsection{Putting All Together}
With the gradient approximations derived above, we can directly plug them into \eqref{eq:wgf_sde} to derive update equations for $\xb$. As $\xb$ is generated from $\zb$ by a pre-trained generator $g$, we are also ready to optimize the latent code $\zb$ by the chain rule\footnote{All the pre-trained DGMs we considered define deterministic mappings from $\zb$ to $\xb$, enabling a direct application of chain rule. One can apply techniques like reparametrization for potentially stochastic mappings.}. Using samples $\zb$ to approximate $q_{\thetab}(\zb|c)$, this corresponds to the generative-model inference task. Specifically, we have
\begin{align}\label{eq:latent_sampling}
        \zb^{t+1} =& \zb^t - \lambda_1 \nabla_{\zb^t} \log q_{\thetab}(g(\zb^t) \vert c) + \lambda_2 \nabla_{\zb^t} \log p(g(\zb^t)) \nonumber \\
        &+  \lambda_3\nabla_{\zb^t} \log p(c \vert g(\zb^t)) 
\end{align}
where $\lambda_1, \lambda_2, \lambda_3$ are non-negative hyper-parameters representing step sizes, and $\zb^0$ is sampled from some initial distribution such as Gaussian. 

Intuitively, the term $-\nabla_{\zb^t} \log q_{\thetab}(g(\zb^t) \vert c)$ increases the generation diversity, as this is the corresponding gradient of maximizing entropy $-\int_{\mathcal{X}} q_{\thetab}(\xb\vert c)\log q_{\thetab}(\xb \vert c) \text{d} \xb$; the term $\nabla_{\zb^t} \log p(g(\zb^t))$ drives the generated distribution to be close to $p(\xb)$, as this is the corresponding gradient of maximizing $\int_{\mathcal{X}} q_{\thetab}(\xb\vert c) \log p(\xb) \text{d} \xb$; finally, the term $\nabla_{\zb^t} \log p(c \vert g(\zb^t))$ forces the generated distribution to match the provided conditions $c$. We can adjust their importance by tuning the hyper-parameters. Generally speaking, all of them try to minimize the KL divergence $D_{KL}(q_{\thetab}(\xb \vert c), p(\xb \vert c))$. Specifically, the first gradient term improves the generation diversity, and the rest improve the generation quality. 

\section{Related Work}
Several recent works \citep{ansari2020refining, che2020your, tanaka2019discriminator, zhou2021learning} can be viewed as relevant to our framework. They aim to 
improve the generation quality of pre-trained models by updating latent vectors with signals from the discriminators.
DOT \citep{tanaka2019discriminator} shows that under certain conditions (Theorem 2 in \citep{tanaka2019discriminator}), the trained discriminator can be used to solve the optimal transport (OT) problem between $q_{\thetab}(\xb)$ and $p(\xb)$. With this, the authors propose an iterative scheme to improve model performance. 
DDLS and DG$f$low \citep{ansari2020refining, che2020your} both try to improve pre-trained GANs based on the assumption $p(\xb) \propto q_{\thetab}(\xb)e^{d(\xb)}$, with $d$ denoting the discriminator of a pre-trained GAN. DDLS tries to minimize the $f$-divergence between the target distribution and the generated distribution, while DG$f$low directly uses Markov chain Monte Carlo (MCMC) with Langevin dynamics to update latent vectors.
By contrast, \citep{zhou2021learning} approximates $q(\xb)$ and $p(\xb)$ using kernel density estimation (KDE) with kernel $k(\xb, \xb^\prime) = \exp(-\Vert d(\xb) - d(\xb^\prime)\Vert^2)$, based on which an iterative scheme to minimize the $f$-divergence is proposed. 

All the aforementioned methods come with limitations.  \citep{ansari2020refining, che2020your, tanaka2019discriminator} are based on assumptions that may not hold in practice, {\it e.g.}, the discriminator is required to be optimal. Furthermore, \citep{che2020your, tanaka2019discriminator} can only be applied to GANs with scalar-valued discriminators; and the density estimation in \citep{zhou2021learning} is not guaranteed to be accurate. All these works are not general enough for multi-modality tasks such as text-to-image generation and text-guided image editing.
On the contrary, our framework does not require impractical assumptions, and can be applied to models with any architecture and objective function. Our density estimation is based on a tractable closed-form solution, which improves over the standard KDE. Furthermore, our framework is generic, including many previous methods \cite{che2020your, zhou2021learning} as its special cases. In addition, our proposed method has been tested on many applications that none of previous methods consider. 


\section{Experiments}
We test our method on different scenarios that correspond to different $c$, including image generation, text-to-image generation, image inpainting, image translation and text-guided image editing.


\begin{table*}[t!]
    \centering
    \begin{sc}
    \begin{adjustbox}{scale=0.8}
    \begin{tabular}{lcccc}
        \toprule
        &\multicolumn{2}{c}{CIFAR-10 $(32 \times 32)$} & \multicolumn{2}{c}{STL-10 $(48 \times 48)$}\\
        Method & FID $(\downarrow)$ & IS $(\uparrow)$ & FID $(\downarrow)$ & IS $(\uparrow)$\\
        \midrule
        WGAN-GP & $28.42 \pm 0.06$ & $7.14 \pm 0.09$ & $39.88 \pm 0.06$ & $8.85 \pm 0.09$ \\
        SN-GAN & $22.30\pm 0.12$ & $7.54 \pm 0.09$ & $40.45 \pm 0.17$ & $8.45 \pm 0.08$\\
        Improved MMD-GAN & $16.21$ & $8.29$ & $36.67$ & $9.36$ \\
        TransGAN & $11.89$ & $ 8.63 \pm 0.16$ & $25.32$ & $10.10 \pm 0.17$\\
        StyleGAN2 + ADA  & $2.42 \pm 0.04$ & $10.14 \pm 0.09$  & -- & --  \\
        \midrule
        WGAN-GP + Ours & $19.21 \pm 0.09$ & $7.89 \pm 0.08$ & $35.50  \pm 0.12$ & $9.33 \pm 0.09$\\
        SN-GAN + Ours & $13.37 \pm 0.08$ & $9.03 \pm 0.10$ & $32.09 \pm 0.08$ & $9.73 \pm 0.14$ \\
        Improved MMD-GAN + Ours & $10.47 \pm 0.04$ & $8.81 \pm 0.11$ & $23.38 \pm 0.03$ & $10.79 \pm 0.07$\\
        TransGAN + Ours & $10.44 \pm 0.07$ & $9.16 \pm 0.16$ & $\mathbf{19.99 \pm 0.04}$ & $\mathbf{11.06 \pm 0.17}$\\
        StyleGAN2 + ADA + Ours & $\mathbf{2.36 \pm 0.02}$ & $\mathbf{10.23 \pm 0.10}$ &--&--\\
        \bottomrule
    \end{tabular}
    \end{adjustbox}
    \end{sc}
    \caption{Improving image generation of pre-trained GANs.}
    \label{tab:image_generation}
\end{table*}
\subsection{Image Generation}
Following previous works \citep{miyato2018spectral}, we apply our framework to improve pre-trained models on CIFAR-10 \citep{krizhevsky2009learning} and STL-10 \citep{coates2011analysis}. The main results in terms of Fr\'echet inception distance (FID) and Inception Score (IS) are reported in Table \ref{tab:image_generation}, which show that our framework can improve the performance of both CNN-based and transformer-based GANs with arbitrary training objectives, achieving the best scores.

Next, to investigate the effectiveness of our gradient approximation by KRR with regularized kernel $k_{\psi}$, we conduct ablation study to compare it with other baselines on the stable SN-GAN model. The results are reported  in Table \ref{tab:ablation_study}. Note DDLS \citep{che2020your} only reports the results on a SN-ResNet-GAN architecture, which is more complex than others models. \citep{li2018gradient} estimates the gradient $\nabla_{\xb} \log q(\xb)$ via Stein's identity \citep{liu2016kernelized}, which unfortunately is not able to estimate $\nabla_{\xb} \log p(\xb)$. To compare with Stein method, we only replace the $\nabla_{\xb} \log q(\xb)$ term in our algorithm with solution from \citep{li2018gradient}.
The results are shown in Table \ref{tab:ablation_study}. Comparing with the simple KDE, our method indeed improves the performance. Furthermore, the smooth kernel $k_{\psi}$ is also seen to significantly improve over the non-smooth version. We also provide more results of ablation study in the Appendix to better understand the impact of each term in \eqref{eq:latent_sampling}. 
\begin{figure*}[t!]
    \centering
    \subfigure[French fries in a red box on the desk.]{
    \includegraphics[width=0.159\linewidth]{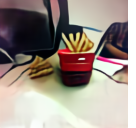}\hfill
    \includegraphics[width=0.149\linewidth]{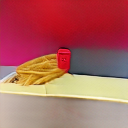}
    }
    \subfigure[A red sphere in a microwave oven.]{
    \includegraphics[width=0.159\linewidth]{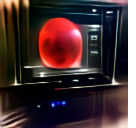}\hfill
    \includegraphics[width=0.149\linewidth]{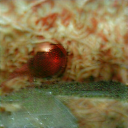}
    }
    \subfigure[A neon sign that reads "VAE".]{
    \includegraphics[width=0.159\linewidth]{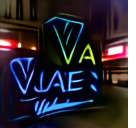}
    \includegraphics[width=0.149\linewidth]{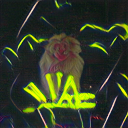}
    }
    \caption{The pre-trained VQ-VAE (left plot of each text input) can handle out-of-domain generation, while BigGAN (right plot of each text input) fails.}
    \label{fig:VQ_VAE_CLIP}
\end{figure*}

\begin{table}[t!]
    \centering
    \begin{sc}
    \begin{adjustbox}{scale=0.75}
    \begin{tabular}{lcccc}
    \toprule
    &\multicolumn{2}{c}{CIFAR-10 $(32 \times 32)$} & \multicolumn{2}{c}{STL-10 $(48 \times 48)$ }\\
    Method & FID $(\downarrow)$ & IS $(\uparrow)$ & FID $(\downarrow)$ & IS $(\uparrow)$\\
    \midrule
     SN-GAN & $22.30\pm 0.12$ & $7.54 \pm 0.09$ & $40.45 \pm 0.17$ & $8.45 \pm 0.08$\\
     DOT & $15.78$ & $8.02 \pm 0.16$ & $34.84$ & $9.35 \pm 0.12$\\
     DDLS & $15.76 $ & $9.05 \pm 0.11$ & -- & --\\
     DG$f$low&  $15.30 \pm 0.08 $ & $8.14 \pm 0.03$ & $34.60 \pm 0.11$ &  $9.66 \pm 0.01$  \\
     KDE & $14.99 \pm 0.09$ & $8.38 \pm 0.13$ & $34.10 \pm 0.12$ & $9.13 \pm 0.10$\\
     \midrule
     KRR + $k$ & $14.56 \pm 0.07$ & $8.48 \pm 0.09$ & $33.23 \pm 0.13$ & $9.24 \pm 0.12$\\
     KDE + $k_{\psi}$ & $13.66 \pm 0.12$ & $8.72 \pm 0.12$ & $32.45 \pm 0.09$ & $9.47 \pm 0.08$\\
     Ours + Stein & $13.62 \pm 0.05$ & $8.65 \pm 0.11$ & $34.49 \pm 0.12$ & $9.09 \pm 0.12$\\
     \textbf{KRR + $k_{\psi}$} & $\mathbf{13.37 \pm 0.08}$ & $9.03 \pm 0.10$ & $\mathbf{32.09 \pm 0.08}$ & $\mathbf{9.73 \pm 0.14}$ \\
     \bottomrule
    \end{tabular}
    \end{adjustbox}
    \end{sc}
    \caption{Ablation study on image generation.}
    \label{tab:ablation_study}
\end{table}


\subsection{Text-to-Image Generation}
We consider applying our framework to improve both in-domain generation and generic generation (zero-shot generation). 
We first evaluate our framework on the widely used CUB \citep{WahCUB_200_2011}  and COCO \citep{lin2014microsoft}. We choose the pre-trained DF-GAN \citep{tao2021dfgan} as our backbone, whose official pre-trained model and implementation are publicly available online. We actually obtain better results on CUB but worse results on COCO than those reported in \citep{tao2021dfgan} when directly evaluating the pre-trained model. The main results are reported in Table \ref{tab:cub_coco}. All the models are evaluated by 30,000 generated images using randomly sampled text descriptions. As mentioned in previous papers \citep{ramesh2021zero, tao2021dfgan}, IS fails in evaluating the text-to-image generation on the COCO dataset, thus we only report FID on COCO. It is seen our proposed method improves both generation quality (FID) and diversity (IS) of the pre-trained model.

\begin{table}[t!]
    \centering
    \begin{sc}
    \begin{adjustbox}{scale=0.75}
    \begin{tabular}{lccc}
    \toprule
    Method & CUB IS $(\uparrow)$ & CUB FID $(\downarrow)$ & COCO FID $(\downarrow)$\\
    \midrule
    AttnGAN & 4.36 & 23.98 & 35.49 \\
    MirrorGAN  & 4.56 & 18.34 & 34.71 \\
    SD-GAN & 4.67 & -- & -- \\
    DM-GAN  & 4.75 & 16.09 & 32.64 \\
    DF-GAN  & 5.14 & 13.63 & 25.91\\
    \textbf{DF-GAN + Ours} &$\mathbf{5.20}$ & $\mathbf{11.93}$ & $\mathbf{21.53}$\\
     \bottomrule
    \end{tabular}
    \end{adjustbox}
    \end{sc}
    \caption{Text-to-image generation on CUB $(256 \times 256)$ and COCO $(256 \times 256)$.}
    \label{tab:cub_coco}
\end{table}

\begin{figure}[t!]
    \centering
    \subfigure[Baseline]{\includegraphics[width=1.0\linewidth]{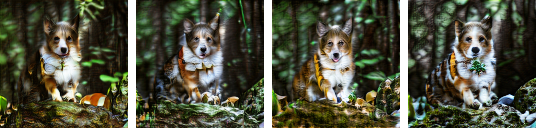}}
    \subfigure[Ours]{\includegraphics[width=1.0\linewidth]{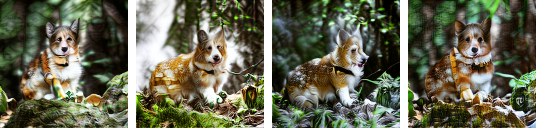}}
    \caption{Text-to-image generation according to 'corgi in the forest' by the pre-trained BigGAN, our method can achieve more diverse generation.}
    \label{fig:biggan_generation}
    \vspace{-0.3cm}
\end{figure}

We also consider more powerful pre-trained models. To this end, we choose the pre-trained BigGAN \citep{brock2018large} as our generator, and use the pre-trained CLIP model \citep{radford2021learning} to estimate the value of $p(c\vert g(\zb))$. 
The image encoder of CLIP is also used to construct the kernel in \eqref{eq:kernel} by applying Gaussian kernel on the extracted image features.
We compare our model with a straightforward baseline, which directly maximizes the text-image similarities evaluated by CLIP w.r.t.\! the latent codes. This corresponds to only considering $\nabla_{\zb} \log p(c\vert g(\zb))$ in our update \eqref{eq:latent_sampling}. Some generated examples are provided in Figure \ref{fig:biggan_generation}, where random seeds are fixed for a better comparison. The baseline method generates very similar images, while ours are much diverse with different background colors and postures. 

We next consider out-of-domain generation (zero-shot generation), which is potentially a more useful scenario in practice. 
To this end, we adopt another powerful generator -- the decoder of the pre-trained VQ-VAE in DALL-E \citep{ramesh2021zero}, which is a powerful text-to-image generation model trained on 250 million text-image pairs. Our framework can seamlessly integrate DALL-E even without its publicly unavailable auto-regressive transformer. 
We achieve this by adopting the CLIP model to estimate the term $\log p(c|g(\zb))$ in \eqref{eq:latent_sampling}.
Some generated samples are provided in Figure \ref{fig:VQ_VAE_CLIP}. Different from BigGAN which has a continuous latent space, the latent space of VQ-VAE is discrete. We thus propose a modified sampling strategy to update the latent codes. Details are provided in the Appendix along with some examples for verification.
    

\subsection{Image Completion/Inpainting}
Image completion (inpainting) can be formulated as a conditional image generation task.
We choose Co-Mod-GAN \citep{zhao2021comodgan} as our baseline model, which is the SOTA method 
for free-form large region image completion. We also report the results of RFR \citep{li2020recurrent}, DeepFillv2 \citep{yu2019free}.
The experiment is implemented at resolution $512 \times 512$ on the FFHQ \citep{karras2019style} and Place2 \citep{zhou2017places}. We follow the mask generation strategies in previous works \citep{yu2019free, zhao2021large}, where the mask ratio is uniformly sampled from $(0.0, 1.0)$. We evaluate the models by FID and Paired/Unpaired-Inception Discriminative Score (P-IDS/U-IDS), where P-IDS indicates the probability that a fake image is considered more realistic than the actual paired real image, and U-IDS indicates the mis-classification rate defined as the average probability that a real images is classified as fake and the vice versa.
The main results are reported in Figure \ref{fig:inpaingting} and Table \ref{tab:image_inpainting}. Some examples are presented in the Appendix. Note \citep{zhao2021large} uses a pre-trained Inception v3 model to extract features from both un-masked real images and completed fake images, then trains a SVM on the extracted features to get P-IDS and U-IDS. 
We notice that P-IDS and U-IDS will be influenced by the regularization hyper-parameter of the SVM, thus we also provide results across different levels of regularization in Figure \ref{fig:inpaingting}. Consistently, our method obtains the better results both quantitatively and qualitatively. 

\subsection{Image Translation}
\begin{table}[t!]
    \centering
    \begin{sc}
    \begin{adjustbox}{scale=0.75}
    \begin{tabular}{lcc}
    \toprule
     & FFHQ $(512 \times 512)$& Places2 $(512 \times 512)$\\
     \midrule
     RFR& $48.7 \pm 0.5$ &  $49.6 \pm 0.2$ \\
     DeepFillv2 & $17.4 \pm 0.4$ &  $22.1 \pm 0.1$\\
     Co-Mod-GAN & $3.7 \pm 0.0$ & $7.9 \pm 0.0$\\
     \textbf{Ours} & $\mathbf{3.6 \pm 0.0}$ & $\mathbf{7.6 \pm 0.0}$\\
     \bottomrule
    \end{tabular}
    \end{adjustbox}
    \end{sc}    
    \caption{FID $(\downarrow)$ of image completion (inpainting).}    \label{tab:image_inpainting}
\end{table}

We conduct image translation experiment on the CelebA-HQ \citep{karras2017progressive} and AFHQ \citep{choi2020stargan} by evaluating the FID score and learned perceptual image patch similarity (LPIPS). Following \citep{choi2020stargan}, we scale all the images to $256 \times 256$, and choose the pre-trained StarGAN v2 \citep{choi2020stargan} as our base model. Following \citep{zhou2021learning}, we use the proposed method to update both style vector $\mathbf{s}$ and latent feature vector $\zb$ in the StarGAN v2. The discriminator of the target domain is used to estimate $p(c\vert \xb)$ and construct the kernel $k_{\psi}$.

The main results are reported in Table \ref{tab:image_translation}, in which latent-guided and reference-guided synthesis are two different implementation proposed in \citep{choi2020stargan}. Compared to MUNIT \citep{huang2018multimodal},  DRIT \citep{lee2018diverse}, MSGAN \citep{mao2019mode}, StarGAN v2 \citep{choi2020stargan}, KDE \citep{zhou2021learning}, our proposed method leads to the best results across different datasets. We notice that there is a trade-off between FID and LPIPS in our proposed method, which is illustrated in Figure \ref{fig:img_translation_fid_lpips} in the Appendix. We test a variety of hyper-parameter settings, report the best FID and the best LPIPS from all the results that are better than the baseline model.

\begin{figure*}[ht!]
    \centering
    \subfigure[Input]{\includegraphics[width=0.10\linewidth]{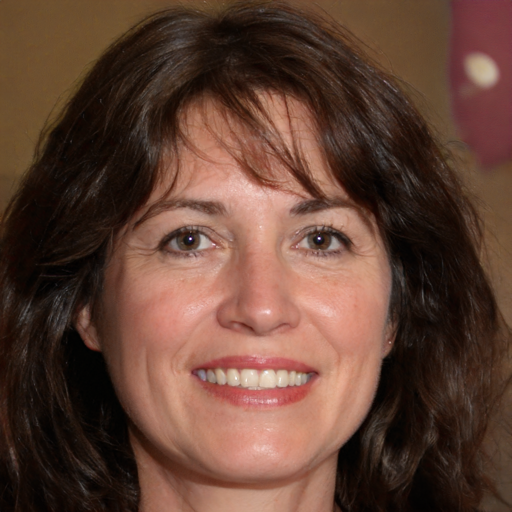}}
    \subfigure["face with straight hair", StyleCLIP-G]{\includegraphics[width=0.4\linewidth]{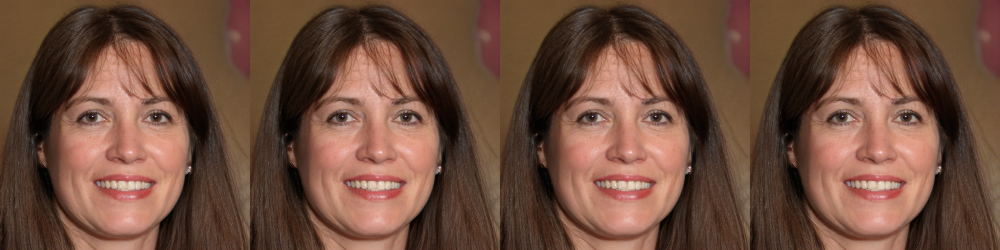}}
    \subfigure["face with straight hair", ours]{\includegraphics[width=0.4\linewidth]{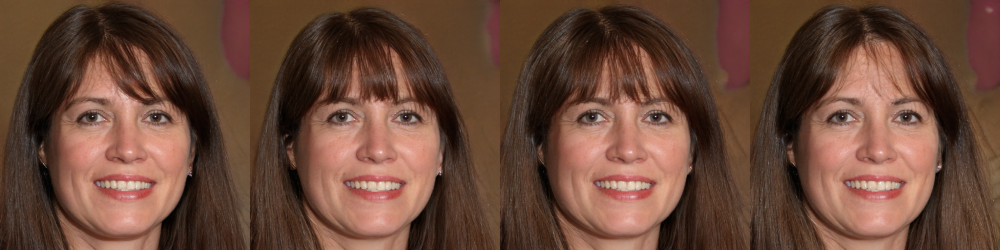}}
    \subfigure[Input]{\includegraphics[width=0.10\linewidth]{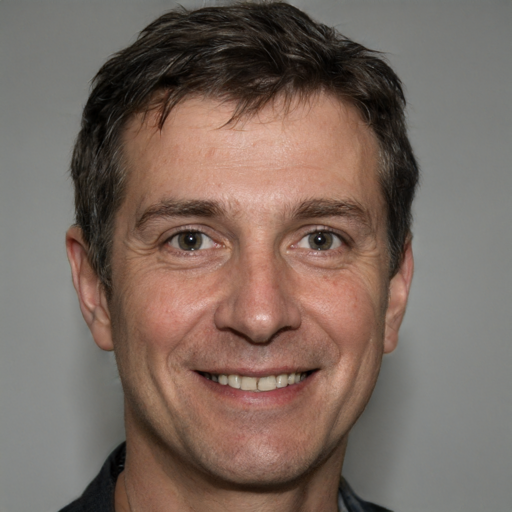}}
    \subfigure["face with beard", StyleCLIP-G]{\includegraphics[width=0.4\linewidth]{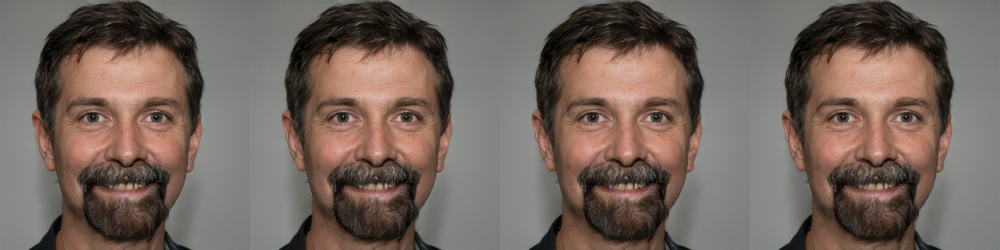}}
    \subfigure["face with beard", ours]{\includegraphics[width=0.4\linewidth]{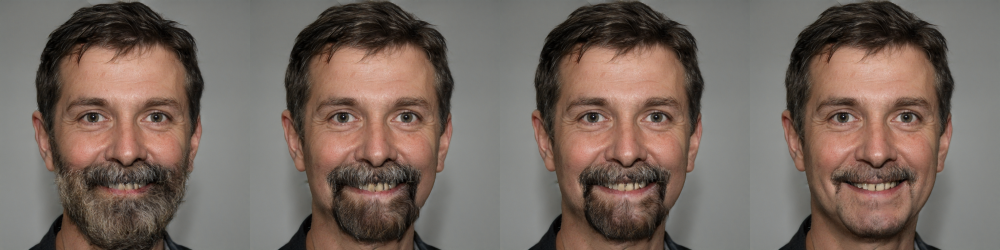}}
    \vspace{-0.1in}
    \caption{Text-guided image editing ($1024\times1024$). StyleCLIP-G cannot generate diverse outputs.}
    \label{fig:img_editing}
\end{figure*}

\begin{figure*}[t!]
    \centering
    \includegraphics[width=0.8\linewidth]{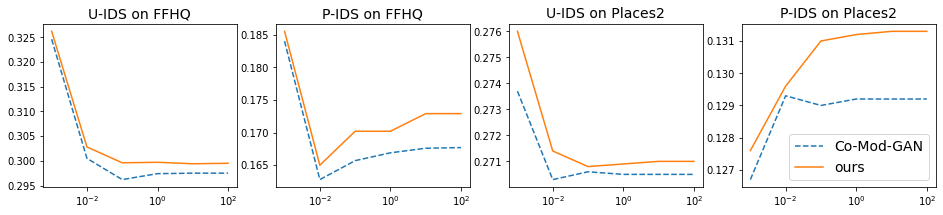}
    \vspace{-0.1in}
    \caption{P-IDS/U-IDS on FFHQ and Place2 datasets.}
    \label{fig:inpaingting}
\end{figure*}

\begin{table*}[ht!]
    \centering
    \begin{sc}
    \begin{adjustbox}{scale=0.8}
    \begin{tabular}{lcccccccc}
        \toprule
        &\multicolumn{4}{c}{CelebA-HQ} & \multicolumn{4}{c}{AFHQ}\\
         Method & \multicolumn{2}{c}{Latent-guided} & \multicolumn{2}{c}{Reference-guided}& \multicolumn{2}{c}{Latent-guided} & \multicolumn{2}{c}{Reference-guided}\\
         & FID $(\downarrow)$ & LPIPS $(\uparrow)$ & FID $(\downarrow)$ & LPIPS $(\uparrow)$& FID $(\downarrow)$ & LPIPS $(\uparrow)$& FID $(\downarrow)$ & LPIPS $(\uparrow)$\\
        \midrule 
        MUNIT & $31.4$ & $0.363$ & $107.1$ & $0.176$ & $41.5$ & $0.511$ & $223.9$ & $0.199$\\
        DRIT  & $52.1$ & $0.178$ & $53.3$ & $0.311$ & $95.6$ & $0.326$ & $114.8$ & $0.156$\\
        MSGAN & $33.1$ & $0.389$ & $39.6$ & $0.312$ & $61.4$ & $0.517$ & $69.8$ & $0.375$\\
        StarGAN v2 & $13.7$ & $0.452$ & $23.8$ & $0.388$ & $16.2$ & $0.450$ & $19.8$ & $0.432$\\
        KDE & $13.3$ & $0.480$ & $17.9$ & $0.488$ & $15.8$ & $0.472$ & $19.2$ & $0.457$\\
        Ours & $\mathbf{13.0}$ & $\mathbf{0.497}$ & $\mathbf{17.4}$ & $\mathbf{0.544}$& $\mathbf{15.6}$ & $0.490$ &$\mathbf{19.0}$ & $\mathbf{0.470}$\\
        \bottomrule
    \end{tabular}
    \end{adjustbox}
    \end{sc}
    \vspace{-0.1in}
    \caption{Improving image translation of pre-trained GANs.}
    \label{tab:image_translation}
\end{table*}

\subsection{Text-guided Image Editing}
Text-guided image editing can be formulated as an image generation task conditioned on both a given text description and an initial image. The recent StyleCLIP \citep{patashnik2021styleclip} solves this problem by combining the pre-trained StyleGAN2 \citep{karras2020analyzing} and CLIP. 
Given images and text guidance, there are three different methods proposed in \citep{patashnik2021styleclip}, including: 1) directly optimizing latent vectors by maximizing the text-image similarity evaluated by pre-trained CLIP; 2) training a mapping from initial latent vector to the revised latent vector for every input text; 3) using CLIP to infer an input-agnostic direction to update the intermediate features of the generator. We choose the last one (denoted as StyleCLIP-G) as our baseline because it leads to the most promising results. However, the resulting images of StyleCLIP-G lack stochasticity because the update direction is input-agnostic and has no randomness. With our proposed method, we expect the resulting images would be more diverse without deteriorating image quality. 
To this end, we treat the update direction obtained from StyleCLIP-G as $\nabla \log p(c \vert \xb)$ in our framework, and construct the kernel by applying Gaussian kernel on the features extracted from the image encoder of pre-trained CLIP.
Some results are presented in Figure \ref{fig:img_editing}, where we can find that our proposed method can generate more diverse images, which are all consistent with the text. More results are provided in the Appendix.

\section{Conclusion}
We propose an effective and generic framework called generative-model inference for better generation  across different scenarios. Specifically, we propose a principled method based on the WGF framework and a regularized kernel density estimator with pre-conditioning weights to solve the WGF. Extensive experiments on a large variety of generation tasks reveal the effectiveness and superiority of the proposed method. We believe our generative-model inference is an exciting research direction that can stimulate further research due to the flexibility of reusing pre-trained SOTA models.

\clearpage
\bibliography{aaai22}

\begin{thebibliography}{46}
\providecommand{\natexlab}[1]{#1}

\bibitem[{Ambrosio, Gigli, and Savar{\'e}(2008)}]{ambrosio2008gradient}
Ambrosio, L.; Gigli, N.; and Savar{\'e}, G. 2008.
\newblock \emph{Gradient flows: in metric spaces and in the space of
  probability measures}.
\newblock Springer Science \& Business Media.

\bibitem[{Ansari, Ang, and Soh(2020)}]{ansari2020refining}
Ansari, A.~F.; Ang, M.~L.; and Soh, H. 2020.
\newblock Refining Deep Generative Models via Wasserstein Gradient Flows.
\newblock \emph{arXiv preprint arXiv:2012.00780}.

\bibitem[{Brock, Donahue, and Simonyan(2018)}]{brock2018large}
Brock, A.; Donahue, J.; and Simonyan, K. 2018.
\newblock Large Scale GAN Training for High Fidelity Natural Image Synthesis.
\newblock In \emph{International Conference on Learning Representations}.

\bibitem[{Carrillo, Craig, and Patacchini(2019)}]{carrillo2019blob}
Carrillo, J.~A.; Craig, K.; and Patacchini, F.~S. 2019.
\newblock A blob method for diffusion.
\newblock \emph{Calculus of Variations and Partial Differential Equations},
  58(2): 1--53.

\bibitem[{Che et~al.(2020)Che, Zhang, Sohl-Dickstein, Larochelle, Paull, Cao,
  and Bengio}]{che2020your}
Che, T.; Zhang, R.; Sohl-Dickstein, J.; Larochelle, H.; Paull, L.; Cao, Y.; and
  Bengio, Y. 2020.
\newblock Your GAN is Secretly an Energy-based Model and You Should use
  Discriminator Driven Latent Sampling.
\newblock \emph{arXiv preprint arXiv:2003.06060}.

\bibitem[{Choi et~al.(2020)Choi, Uh, Yoo, and Ha}]{choi2020stargan}
Choi, Y.; Uh, Y.; Yoo, J.; and Ha, J.-W. 2020.
\newblock Stargan v2: Diverse image synthesis for multiple domains.
\newblock In \emph{Proceedings of the IEEE/CVF Conference on Computer Vision
  and Pattern Recognition}, 8188--8197.

\bibitem[{Coates, Ng, and Lee(2011)}]{coates2011analysis}
Coates, A.; Ng, A.; and Lee, H. 2011.
\newblock An analysis of single-layer networks in unsupervised feature
  learning.
\newblock In \emph{Proceedings of the fourteenth international conference on
  artificial intelligence and statistics}, 215--223. JMLR Workshop and
  Conference Proceedings.

\bibitem[{Craig and Bertozzi(2016)}]{craig2016blob}
Craig, K.; and Bertozzi, A. 2016.
\newblock A blob method for the aggregation equation.
\newblock \emph{Mathematics of computation}, 85(300): 1681--1717.

\bibitem[{Gao et~al.(2020)Gao, Chen, Liu, Tan, and Yan}]{Gao_2020_CVPR}
Gao, C.; Chen, Y.; Liu, S.; Tan, Z.; and Yan, S. 2020.
\newblock AdversarialNAS: Adversarial Neural Architecture Search for GANs.
\newblock In \emph{Proceedings of the IEEE/CVF Conference on Computer Vision
  and Pattern Recognition (CVPR)}.

\bibitem[{Gong et~al.(2019)Gong, Chang, Jiang, and Wang}]{gong2019autogan}
Gong, X.; Chang, S.; Jiang, Y.; and Wang, Z. 2019.
\newblock Autogan: Neural architecture search for generative adversarial
  networks.
\newblock In \emph{Proceedings of the IEEE/CVF International Conference on
  Computer Vision}, 3224--3234.

\bibitem[{Goodfellow et~al.(2014)Goodfellow, Pouget-Abadie, Mirza, Xu,
  Warde-Farley, Ozair, Courville, and Bengio}]{goodfellow2014generative}
Goodfellow, I.~J.; Pouget-Abadie, J.; Mirza, M.; Xu, B.; Warde-Farley, D.;
  Ozair, S.; Courville, A.~C.; and Bengio, Y. 2014.
\newblock Generative Adversarial Nets.
\newblock In \emph{NIPS}.

\bibitem[{Ho, Jain, and Abbeel(2020)}]{NEURIPS2020_4c5bcfec}
Ho, J.; Jain, A.; and Abbeel, P. 2020.
\newblock Denoising Diffusion Probabilistic Models.
\newblock In Larochelle, H.; Ranzato, M.; Hadsell, R.; Balcan, M.~F.; and Lin,
  H., eds., \emph{Advances in Neural Information Processing Systems},
  volume~33, 6840--6851. Curran Associates, Inc.

\bibitem[{Huang et~al.(2018)Huang, Liu, Belongie, and
  Kautz}]{huang2018multimodal}
Huang, X.; Liu, M.-Y.; Belongie, S.; and Kautz, J. 2018.
\newblock Multimodal unsupervised image-to-image translation.
\newblock In \emph{Proceedings of the European conference on computer vision
  (ECCV)}, 172--189.

\bibitem[{It{\^o}(1951)}]{ito1951stochastic}
It{\^o}, K. 1951.
\newblock \emph{On stochastic differential equations}.
\newblock 4. American Mathematical Soc.

\bibitem[{Karras et~al.(2017)Karras, Aila, Laine, and
  Lehtinen}]{karras2017progressive}
Karras, T.; Aila, T.; Laine, S.; and Lehtinen, J. 2017.
\newblock Progressive growing of gans for improved quality, stability, and
  variation.
\newblock \emph{arXiv preprint arXiv:1710.10196}.

\bibitem[{Karras, Laine, and Aila(2019)}]{karras2019style}
Karras, T.; Laine, S.; and Aila, T. 2019.
\newblock A style-based generator architecture for generative adversarial
  networks.
\newblock In \emph{Proceedings of the IEEE/CVF Conference on Computer Vision
  and Pattern Recognition}, 4401--4410.

\bibitem[{Karras et~al.(2020)Karras, Laine, Aittala, Hellsten, Lehtinen, and
  Aila}]{karras2020analyzing}
Karras, T.; Laine, S.; Aittala, M.; Hellsten, J.; Lehtinen, J.; and Aila, T.
  2020.
\newblock Analyzing and improving the image quality of stylegan.
\newblock In \emph{Proceedings of the IEEE/CVF Conference on Computer Vision
  and Pattern Recognition}, 8110--8119.

\bibitem[{Krizhevsky, Hinton et~al.(2009)}]{krizhevsky2009learning}
Krizhevsky, A.; Hinton, G.; et~al. 2009.
\newblock Learning multiple layers of features from tiny images.

\bibitem[{Lee et~al.(2018)Lee, Tseng, Huang, Singh, and Yang}]{lee2018diverse}
Lee, H.-Y.; Tseng, H.-Y.; Huang, J.-B.; Singh, M.; and Yang, M.-H. 2018.
\newblock Diverse image-to-image translation via disentangled representations.
\newblock In \emph{Proceedings of the European conference on computer vision
  (ECCV)}, 35--51.

\bibitem[{Li et~al.(2020)Li, Wang, Zhang, Du, and Tao}]{li2020recurrent}
Li, J.; Wang, N.; Zhang, L.; Du, B.; and Tao, D. 2020.
\newblock Recurrent feature reasoning for image inpainting.
\newblock In \emph{Proceedings of the IEEE/CVF Conference on Computer Vision
  and Pattern Recognition}, 7760--7768.

\bibitem[{Li and Turner(2018)}]{li2018gradient}
Li, Y.; and Turner, R.~E. 2018.
\newblock Gradient Estimators for Implicit Models.
\newblock In \emph{International Conference on Learning Representations}.

\bibitem[{Lin et~al.(2014)Lin, Maire, Belongie, Hays, Perona, Ramanan,
  Doll{\'a}r, and Zitnick}]{lin2014microsoft}
Lin, T.-Y.; Maire, M.; Belongie, S.; Hays, J.; Perona, P.; Ramanan, D.;
  Doll{\'a}r, P.; and Zitnick, C.~L. 2014.
\newblock Microsoft coco: Common objects in context.
\newblock In \emph{European conference on computer vision}, 740--755. Springer.

\bibitem[{Liu, Lee, and Jordan(2016)}]{liu2016kernelized}
Liu, Q.; Lee, J.; and Jordan, M. 2016.
\newblock A kernelized Stein discrepancy for goodness-of-fit tests.
\newblock In \emph{International conference on machine learning}, 276--284.
  PMLR.

\bibitem[{Mao et~al.(2019)Mao, Lee, Tseng, Ma, and Yang}]{mao2019mode}
Mao, Q.; Lee, H.-Y.; Tseng, H.-Y.; Ma, S.; and Yang, M.-H. 2019.
\newblock Mode seeking generative adversarial networks for diverse image
  synthesis.
\newblock In \emph{Proceedings of the IEEE/CVF Conference on Computer Vision
  and Pattern Recognition}, 1429--1437.

\bibitem[{Meng et~al.(2021)Meng, Song, Song, Zhao, and
  Ermon}]{meng2021improved}
Meng, C.; Song, J.; Song, Y.; Zhao, S.; and Ermon, S. 2021.
\newblock Improved Autoregressive Modeling with Distribution Smoothing.
\newblock \emph{arXiv preprint arXiv:2103.15089}.

\bibitem[{Miyato et~al.(2018)Miyato, Kataoka, Koyama, and
  Yoshida}]{miyato2018spectral}
Miyato, T.; Kataoka, T.; Koyama, M.; and Yoshida, Y. 2018.
\newblock Spectral Normalization for Generative Adversarial Networks.
\newblock In \emph{International Conference on Learning Representations}.

\bibitem[{Nguyen et~al.(2017)Nguyen, Clune, Bengio, Dosovitskiy, and
  Yosinski}]{nguyen2017plug}
Nguyen, A.; Clune, J.; Bengio, Y.; Dosovitskiy, A.; and Yosinski, J. 2017.
\newblock Plug \& play generative networks: Conditional iterative generation of
  images in latent space.
\newblock In \emph{Proceedings of the IEEE Conference on Computer Vision and
  Pattern Recognition}, 4467--4477.

\bibitem[{Patashnik et~al.(2021)Patashnik, Wu, Shechtman, Cohen-Or, and
  Lischinski}]{patashnik2021styleclip}
Patashnik, O.; Wu, Z.; Shechtman, E.; Cohen-Or, D.; and Lischinski, D. 2021.
\newblock StyleCLIP: Text-Driven Manipulation of StyleGAN Imagery.
\newblock \emph{arXiv preprint arXiv:2103.17249}.

\bibitem[{Qiao et~al.(2019)Qiao, Zhang, Xu, and Tao}]{qiao2019mirrorgan}
Qiao, T.; Zhang, J.; Xu, D.; and Tao, D. 2019.
\newblock Mirrorgan: Learning text-to-image generation by redescription.
\newblock In \emph{Proceedings of the IEEE/CVF Conference on Computer Vision
  and Pattern Recognition}, 1505--1514.

\bibitem[{Radford et~al.(2021)Radford, Kim, Hallacy, Ramesh, Goh, Agarwal,
  Sastry, Askell, Mishkin, Clark, Krueger, and Sutskever}]{radford2021learning}
Radford, A.; Kim, J.~W.; Hallacy, C.; Ramesh, A.; Goh, G.; Agarwal, S.; Sastry,
  G.; Askell, A.; Mishkin, P.; Clark, J.; Krueger, G.; and Sutskever, I. 2021.
\newblock Learning Transferable Visual Models From Natural Language
  Supervision.
\newblock arXiv:2103.00020.

\bibitem[{Ramesh et~al.(2021)Ramesh, Pavlov, Goh, Gray, Voss, Radford, Chen,
  and Sutskever}]{ramesh2021zero}
Ramesh, A.; Pavlov, M.; Goh, G.; Gray, S.; Voss, C.; Radford, A.; Chen, M.; and
  Sutskever, I. 2021.
\newblock Zero-shot text-to-image generation.
\newblock \emph{arXiv preprint arXiv:2102.12092}.

\bibitem[{Risken(1996)}]{risken1996fokker}
Risken, H. 1996.
\newblock Fokker-planck equation.
\newblock In \emph{The Fokker-Planck Equation}, 63--95. Springer.

\bibitem[{Santambrogio(2016)}]{santambrogio2016}
Santambrogio, F. 2016.
\newblock { Euclidean, Metric, and Wasserstein } Gradient Flows: an overview.
\newblock arXiv:1609.03890.

\bibitem[{Song, Meng, and Ermon(2021)}]{song2021denoising}
Song, J.; Meng, C.; and Ermon, S. 2021.
\newblock Denoising Diffusion Implicit Models.
\newblock In \emph{International Conference on Learning Representations}.

\bibitem[{Song et~al.(2021)Song, Sohl-Dickstein, Kingma, Kumar, Ermon, and
  Poole}]{song2021scorebased}
Song, Y.; Sohl-Dickstein, J.; Kingma, D.~P.; Kumar, A.; Ermon, S.; and Poole,
  B. 2021.
\newblock Score-Based Generative Modeling through Stochastic Differential
  Equations.
\newblock In \emph{International Conference on Learning Representations}.

\bibitem[{Tanaka(2019)}]{tanaka2019discriminator}
Tanaka, A. 2019.
\newblock Discriminator optimal transport.
\newblock In \emph{Advances in Neural Information Processing Systems},
  6816--6826.

\bibitem[{Tao et~al.(2021)Tao, Tang, Wu, Sebe, Jing, Wu, and
  Bao}]{tao2021dfgan}
Tao, M.; Tang, H.; Wu, S.; Sebe, N.; Jing, X.-Y.; Wu, F.; and Bao, B. 2021.
\newblock DF-GAN: Deep Fusion Generative Adversarial Networks for Text-to-Image
  Synthesis.
\newblock arXiv:2008.05865.

\bibitem[{Tropp(2015)}]{tropp2015introduction}
Tropp, J.~A. 2015.
\newblock An introduction to matrix concentration inequalities.
\newblock \emph{arXiv preprint arXiv:1501.01571}.

\bibitem[{Wah et~al.(2011)Wah, Branson, Welinder, Perona, and
  Belongie}]{WahCUB_200_2011}
Wah, C.; Branson, S.; Welinder, P.; Perona, P.; and Belongie, S. 2011.
\newblock {The Caltech-UCSD Birds-200-2011 Dataset}.
\newblock Technical Report CNS-TR-2011-001, California Institute of Technology.

\bibitem[{Xu et~al.(2018)Xu, Zhang, Huang, Zhang, Gan, Huang, and
  He}]{xu2018attngan}
Xu, T.; Zhang, P.; Huang, Q.; Zhang, H.; Gan, Z.; Huang, X.; and He, X. 2018.
\newblock Attngan: Fine-grained text to image generation with attentional
  generative adversarial networks.
\newblock In \emph{Proceedings of the IEEE conference on computer vision and
  pattern recognition}, 1316--1324.

\bibitem[{Yu et~al.(2019)Yu, Lin, Yang, Shen, Lu, and Huang}]{yu2019free}
Yu, J.; Lin, Z.; Yang, J.; Shen, X.; Lu, X.; and Huang, T.~S. 2019.
\newblock Free-form image inpainting with gated convolution.
\newblock In \emph{Proceedings of the IEEE International Conference on Computer
  Vision}, 4471--4480.

\bibitem[{Zhao et~al.(2021{\natexlab{a}})Zhao, Cui, Sheng, Dong, Liang, Chang,
  and Xu}]{zhao2021comodgan}
Zhao, S.; Cui, J.; Sheng, Y.; Dong, Y.; Liang, X.; Chang, E.~I.; and Xu, Y.
  2021{\natexlab{a}}.
\newblock Large Scale Image Completion via Co-Modulated Generative Adversarial
  Networks.
\newblock In \emph{International Conference on Learning Representations
  (ICLR)}.

\bibitem[{Zhao et~al.(2021{\natexlab{b}})Zhao, Cui, Sheng, Dong, Liang, Chang,
  and Xu}]{zhao2021large}
Zhao, S.; Cui, J.; Sheng, Y.; Dong, Y.; Liang, X.; Chang, E.~I.; and Xu, Y.
  2021{\natexlab{b}}.
\newblock Large Scale Image Completion via Co-Modulated Generative Adversarial
  Networks.
\newblock arXiv:2103.10428.

\bibitem[{Zhou et~al.(2017)Zhou, Lapedriza, Khosla, Oliva, and
  Torralba}]{zhou2017places}
Zhou, B.; Lapedriza, A.; Khosla, A.; Oliva, A.; and Torralba, A. 2017.
\newblock Places: A 10 million Image Database for Scene Recognition.
\newblock \emph{IEEE Transactions on Pattern Analysis and Machine
  Intelligence}.

\bibitem[{Zhou, Chen, and Xu(2021)}]{zhou2021learning}
Zhou, Y.; Chen, C.; and Xu, J. 2021.
\newblock Learning High-Dimensional Distributions with Latent Neural
  Fokker-Planck Kernels.
\newblock arXiv:2105.04538.

\bibitem[{Zhu et~al.(2017)Zhu, Park, Isola, and Efros}]{zhu2017unpaired}
Zhu, J.-Y.; Park, T.; Isola, P.; and Efros, A.~A. 2017.
\newblock Unpaired image-to-image translation using cycle-consistent
  adversarial networks.
\newblock In \emph{Proceedings of the IEEE international conference on computer
  vision}, 2223--2232.

\end{thebibliography}

\onecolumn
\appendix

\section{Proof of Theorem \ref{lemma:wgf_kl}}
\textbf{Theorem \ref{lemma:wgf_kl}}
\textit{
Denote $\Div$ as the divergence operation, and $p(c \vert \xb)$ the likelihood of $c$ given $\xb$. The Wasserstein gradient flow of the functional $\mathcal{F}(q) = \mathcal{D}_{KL}(q_{\thetab}(\xb\vert c), p(\xb \vert c))$ can be represented by by the  partial differential equation (PDE):
    \begin{align}
        \dfrac{\partial q_{\thetab}(\xb\vert c)}{\partial t} = \Div (q_{\thetab}(\xb\vert c) \nabla_{\xb} \log q_{\thetab}(\xb\vert c) - q_{\thetab}(\xb\vert c)\nabla_{\xb} \log p(\xb) - q_{\thetab}(\xb\vert c) \nabla_{\xb} \log p(c \vert \xb)  )
    \end{align}
The associated ordinary differential equation (ODE) for $\xb$ is:
    \begin{align}
        \text{d}\xb = -\nabla_{\xb} \log q_{\thetab}(\xb\vert c) \text{d}t + \nabla_{\xb} \log p(\xb) \text{d}t + \nabla_{\xb} \log p(c \vert \xb) \text{d}t~.
    \end{align}
}
\par\noindent{\bf Proof\ }
\begin{align}
    &\mathcal{D}_{KL}(q_{\thetab}(\xb\vert c), p(\xb \vert c)) \nonumber \\
    =& \int q_{\thetab}(\xb \vert c) \log q_{\thetab}(\xb \vert c) - q_{\thetab}(\xb \vert c) \log p(\xb \vert c)\text{d}\xb \nonumber \\
    =& \int q_{\thetab}(\xb \vert c) \log q_{\thetab}(\xb \vert c) - q_{\thetab}(\xb \vert c) \log \dfrac{p(\xb, c)}{p(c)}\text{d}\xb \nonumber \\
    = & \int q_{\thetab}(\xb \vert c) \log q_{\thetab}(\xb \vert c) - q_{\thetab}(\xb \vert c) \log \dfrac{p(c \vert \xb)p(\xb)}{p(c)}\text{d}\xb \nonumber \\
    = & \int q_{\thetab}(\xb \vert c) \log q_{\thetab}(\xb \vert c) - q_{\thetab}(\xb \vert c) \log p(c \vert \xb) - q_{\thetab}(\xb \vert c) \log p(\xb)\text{d}\xb + A\label{eq:re-write-kl}
\end{align}
where $A$ is an unknown constant.
We know that the WGF minimizing functional $\mathcal{F}(q)$ can be written as $\dfrac{\partial q}{\partial t} = \text{Div} (q \nabla_{\zb} \dfrac{\delta \mathcal{F}(q)}{\delta q})$, where $\dfrac{\delta \mathcal{F}(q)}{\delta q}$ denotes the first variation of the funtional \citep{ambrosio2008gradient, santambrogio2016}. 

Assume that $f: R \rightarrow R$, $V: \Omega \rightarrow R$ and $W: R^d \rightarrow R$ are regular enough, while $W$ is symmetric. Then, the first variation of the following functionals:
\[
\mathcal{F}(q) = \int (f(q)) \text{d}\xb, \mathcal{V}(q) = \int V(\xb) \text{d}q, \mathcal{W}(q) = \dfrac{1}{2}\int \int W(\xb - \yb) \text{d}q(\xb) \text{d} q(\yb)
\]
are
\[
\dfrac{\delta \mathcal{F}}{\delta q}(q) = f^\prime (q), \dfrac{\delta \mathcal{V}}{\delta q}(q) = V, \dfrac{\delta \mathcal{W}}{\delta q}(q) = W * q.
\]
It is easy to get the first variation of \eqref{eq:re-write-kl}, and obtain the PDE for its WGF as:
\[
    \dfrac{\partial q_{\thetab}(\xb\vert c)}{\partial t} = \Div (q_{\thetab}(\xb\vert c) \nabla_{\xb} \log q_{\thetab}(\xb\vert c) - q_{\thetab}(\xb\vert c)\nabla_{\xb} \log p(\xb) - q_{\thetab}(\xb\vert c) \nabla_{\xb} \log p(c \vert \xb)  ))
\]
By \cite{ito1951stochastic, risken1996fokker}, we can easily get the corresponding ODE as
\[
    \text{d}\xb = -\nabla_{\xb} \log q_{\thetab}(\xb\vert c) \text{d}t + \nabla_{\xb} \log p(\xb) \text{d}t + \nabla_{\xb} \log p(c \vert \xb) \text{d}t.
\]
Similar idea can also be applied to arbitrary $f$-divergence:
\[
\mathcal{D}_f(q_{\thetab}(\xb\vert c), p(\xb \vert c)) = \int q_{\thetab}(\xb \vert c) f(p(\xb \vert c)/q_{\thetab}(\xb \vert c)),
\]
in which we also use Bayes rule to re-write the unknown distribution $p(\xb \vert c)$.

\hfill\BlackBox\\[2mm]

\section{Proof of Theorem \ref{thm:bounded_error}}

\textbf{Theorem \ref{thm:bounded_error}}
\textit{Assume $k_{\xb}$ is twice differential with bounded derivatives, $\psi(\epsilonb)$ denotes a distribution with zero mean and bounded variance $\sigma^2 < \infty$. Let $k_{\psi}(\xb, \xb^\prime) = (k_{\xb} * \psi)(\xb^\prime)$ be the regularized kernel, then
$
\vert (k_{\xb} * \psi)(\xb^\prime) - k_{\xb}(\xb^\prime) \vert \leq  O(\sigma^2).
$}
\par\noindent{\bf Proof\ }
By Taylor expansion, we have
\[
k_{\xb}(\xb^\prime - \epsilonb) \leq k_{\xb}(\xb^\prime) - \epsilonb \nabla k_{\xb}(\xb^\prime) + O(\epsilonb^2)
\]
integrate with $\psi(\epsilonb)$, we have
\[
\int k_{\xb}(\xb^\prime - \epsilonb) \psi(\epsilonb) \text{d} \epsilonb \leq \int k_{\xb}(\xb^\prime) \psi(\epsilonb) \text{d} \epsilonb - \int \epsilonb \nabla k_{\xb}(\xb^\prime) \psi(\epsilonb) \text{d} \epsilonb + \int C \epsilonb^2 \psi(\epsilonb) \text{d} \epsilonb
\]
where $C$ is an unknown constant. By Previous Lemma, we know that $\psi(\epsilonb) $ is actually PDF of distribution with zero mean and bounded variance, thus we have
\[
(k_{\xb}*\psi)(\xb^\prime) \leq k_{\xb}(\xb^\prime) - \nabla k_{\xb}(\xb^\prime) \int \epsilonb  \psi(\epsilonb) \text{d} \epsilonb + C_1 \sigma^2
\]
Thus we have
\[
(k_{\xb}*\psi)(\xb^\prime) - k_{\xb}(\xb^\prime) \leq  C_1 \sigma^2
\]
Similarly, we have
\[
 k_{\xb}(\xb^\prime) - (k_{\xb}*\psi)(\xb^\prime) \leq  C_2 \sigma^2
\]
Thus
\[
\vert (k_{\xb} * \psi)(\xb^\prime) - k_{\xb}(\xb^\prime) \vert \leq  O(\sigma^2)
\]
\hfill\BlackBox\\[2mm]

\section{Proof of Theorem \ref{thm:krr_kde}}
\textbf{Theorem \ref{thm:krr_kde}}
\textit{
Let $K$ be a $n \times n$ kernel matrix, $\eta>0$. Then the following holds: $\Vert (\eta I_{n} + K)^{-1} - \eta^{-1} I_{n}\Vert^2_{\mathcal{F}} \leq O(n\eta^{-4})$, where $\Vert \cdot \Vert_{\mathcal{F}}$ denotes the Frobenius norm of matrix. In other words, when increasing the regularization level $\eta$, our proposed method approaches to the kernel density estimator.
}
\par\noindent{\bf Proof\ }
Re-write
\begin{align*}
    &\Vert (\eta I_{n} + K)^{-1} - \eta^{-1} I_{n}\Vert^2_{\mathcal{F}}\\
    =& \eta^{-2} \Vert \eta (\eta I_{n} + K)^{-1} -  I_{n}\Vert^2_{\mathcal{F}}.
\end{align*}
Because
\[
\eta (\eta I_n + K)^{-1} (\eta I_n + K) \eta^{-1} = I_n,
\]
we have
\[
\eta (\eta I_n + K)^{-1} + (\eta I_n + K)^{-1}K = I_n.
\]
\[
\eta (\eta I_n + K)^{-1} = I_n - (\eta I_n + K)^{-1}K 
\]
Substitute the above into previous equation, we have
\begin{align*}
    &\Vert (\eta I_{n} + K)^{-1} - \eta^{-1} I_{n}\Vert^2_{\mathcal{F}}\\
    =& \eta^{-2} \Vert \eta (\eta I_{n} + K)^{-1} -  I_{n}\Vert^2_{\mathcal{F}}\\
    =& \eta^{-2} \Vert (\eta I_n + K)^{-1}K\Vert^2_{\mathcal{F}} \\
    \leq & \eta^{-2} \Vert (\eta I_n + K)^{-1}\Vert^2_{\mathcal{F}}\Vert K\Vert^2_{\mathcal{F}}\\
    \leq & n\eta^{-2} \Vert (\eta I_n + K)^{-1}\Vert^2_2\Vert K\Vert^2_{\mathcal{F}}\\
\end{align*}
where $\Vert \cdot \Vert_2$ denotes the spectral norm of the matrix, the last inequality comes from the relation between Frobenius norm and spectral norm:
\[
\Vert A \Vert_2 \leq \Vert A\Vert_{\mathcal{F}} \leq \sqrt{r} \Vert A \Vert _2
\]
where $r$ is the rank of the matrix.

Because $K$ is the kernel matrix of positive semi-definite kernel by construction, both $\eta I_n$ and $K$ are positive semi-definite matrices. We know that
\[
\eta I_n \preccurlyeq \eta I_n +K
\]
where $\preccurlyeq$ denotes the semi-definite partial order, $A \preccurlyeq B$ if and only if $B-A$ is a positive semi-definite matrix.
By Proposition 8.4.3 in \citep{tropp2015introduction}, we know that if $\eta I_n \preccurlyeq \eta I_n + K$, then
\[
(\eta I + K)^{-1} \preccurlyeq (\eta I_n)^{-1}.
\]
Consequently, we have
\[
(\lambda_{min}(\eta I_n + K))^{-2} \leq (\lambda_{min}(\eta I_n))^{-2}
\]
from the Fact 8.3.2 in \citep{tropp2015introduction}, where $\lambda_{min}(A)$ denotes the minimum eigenvalue of matrix $A$.

Recall that both $\eta I_n$ and $K$ are Hermitian matrices, they are also both positive semi-definite, which means their eigenvalues are all non-negative. Then we can know that
\begin{align*}
    &\Vert (\eta I_{n} + K)^{-1} - \eta^{-1} I_{n}\Vert^2_{\mathcal{F}}\\
    \leq & n\eta^{-2} \Vert(\eta I_n + K)^{-1}\Vert^2_2  \Vert K\Vert^2_{\mathcal{F}}\\
    \leq & n\eta^{-2} (\lambda_{max}((\eta I_n + K)^{-1}))^2  \Vert K\Vert^2_{\mathcal{F}}\\
    \leq & n\eta^{-2} (\lambda_{min}(\eta I_n + K))^{-2}  \Vert K\Vert^2_{\mathcal{F}}\\
    \leq & n\eta^{-2} (\lambda_{min}(\eta I_n))^{-2}\\
    \leq & O(n\eta^{-4})
\end{align*}
which completes the proof.

\hfill\BlackBox\\[2mm]

\section{Discussion on Regularized Kernel}\label{app:regularizedkernel}
In our proposed method, we introduce a mollifier function to regularize the kernel by convolution, which is also used in some previous works \citep{craig2016blob, carrillo2019blob}. As claimed in previous works, the mollifier function will lead to a stable particle based solution. Here we briefly discuss why it leads to better results in our framework.

It is often assumed that data samples actually lie on a low-dimensional manifold embedded in the high-dimensional data space. As a result, the probability density functions may have very sharp transitions around the boundary of the manifold, which means the kernels constructed using pre-trained networks may not be able to model the unknown distribution well.

In \citep{meng2021improved}, the authors first model a smoothed version of the target distribution, then they reverse the smoothing procedure to model the real target distribution, which can also be understood as denoising. 
In our proposed method, regularizing the kernel leads to a regularized or smoothed version of the target distribution, which is easier to model/sample from. In practice, we may use a sequence of mollifier function, and gradually decrease their variance. When $\sigma = 0$, we are actually modeling the real target distribution. 

\section{Details on Text-to-image Generation}

\begin{figure}[h!]
    \centering
    \subfigure[]{\includegraphics[width=0.15\linewidth]{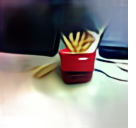}}
    \subfigure[]{\includegraphics[width=0.15\linewidth]{figures/VQ-VAE-CLIP/French_fries_in_a_red_box_on_the_desk.9.seed.1.0.005.0.0.0.0.100.64.0.0.png}}
    \subfigure[]{\includegraphics[width=0.15\linewidth]{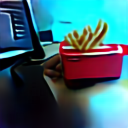}}
    \caption{Generation according to "French fries in a red box on the desk" for VQ-VAE. (a). Generated image after the warm-up stage; (b).Generated image after the fine-tune stage; (c).Generated image without regularization.}
    \label{fig:text_to_img_generation_vq_vae_example}
\end{figure}

We conduct most of the experiments on one Nvidia RTX 2080 Ti GPU whenever applicable, one Nvidia RTX 3090 is used when larger memory is needed (e.g. high-resolution case).

In the experiment of text-to-image generation with VQ-VAE, we update $\zb$ in a different way with improving pre-trained GANs. 
Due to the fact that VQ-VAE's latent space is actually discrete, our proposed method can not be directly applied as it happens in continous space.

Denote the latent code as $\zb = \left[\zb_1, \zb_2, ..., \zb_k\right]$, which is a collection of $k$ features.
We propose to update the latent code in continuous space, while adding a regularizer:
\[
\alpha \sum_{i=1}^k\Vert \zb_i - \eb_i\Vert ^2
\]
where $\eb_i$ is the closest feature to $\zb_i$ in the dictionary, $\alpha>0$ is a hyper-parameter. 

We adopt a two-stage scheme: we first update all $k$ features for several steps, then autoregressively update each of $k$ latent feature using some pre-defined order. Intuitively, the first stage corresponds to a warm-up stage, which provide a global structure of the generated image. The second stage corresponds to a fine-tune stage, which optimize the details of the generated image. 

An illustration is provided in Figure \ref{fig:text_to_img_generation_vq_vae_example}. From which we can find that regularization indeed improves the generation quality, and the two stage scheme works as we expected.




\section{More Experimental Results} 
\paragraph{Image Generation}
We provide some more results on image generation here. First of all, we investigate the impacts of step size and step number of samplings, whose results are shown in Table \ref{tab:diff_step}. The step sizes are normally chosen based on hyper-parameter tuning. They are typically in the range [0.1, 1] for most experiments. Small step sizes will need more sampling steps, but they are guaranteed to obtain results better than baselines (pre-trained GANs without sampling). Using larger step sizes may be efficient, as it only need a few updates to get reasonable results. But when the step sizes are too large, the proposed method may fail to improve the pre-trained models due to large numerical errors.

\begin{table}[h!]
    \centering
    \begin{tabular}{lcccccc}
        \toprule
        step number /\ step size	& 0.1 & 0.2&0.3&0.5&1.0&2.0\\
        \midrule 
        5 steps& $16.88 \pm 0.10$ & $14.81 \pm 0.06$ & $13.71 \pm 0.11$ & $14.97 \pm 0.09$ & $17.67 \pm 0.09$ & $20.45 \pm 0.09$ \\			
        10 steps& $16.22 \pm 0.03$ & $14.47 \pm 0.11$ &$13.52 \pm 0.08$ & $14.48 \pm 0.07$ & $17.36 \pm 0.13$ & $20.30\pm0.08$\\
        15 steps& $16.28 \pm 0.10$ & $14.21 \pm 0.05$ & $13.37 \pm 0.09$ & $14.33 \pm 0.08$ & $17.14 \pm 0.12$ & $20.07 \pm 0.05$\\
         \bottomrule
    \end{tabular}
     \caption{Results of different step sizes and step numbers on CIFAR-10 dataset.}\label{tab:diff_step}
\end{table}

We then investigate the impact of of each term in \eqref{eq:latent_sampling}, the results are provided in  Table \ref{tab:diff_term}. We set the step size to be 0.3, sampling steps to be 10.

\begin{table}[h!]
    \centering
    \begin{tabular}{lccc}
    \toprule
        Method & FID $\downarrow$ & IS $\uparrow$\\
        \midrule
         SN-GAN (baseline) & $22.30 \pm 0.12$ & $7.54 \pm 0.09$ \\
         sampling with only $p(x)$ & $13.76 \pm 0.15$ & $8.62 \pm 0.09$ \\
         sampling with only $q(x\vert c)$ & $18.90 \pm 0.15$ & $7.92 \pm 0.10$ \\
          sampling with only $p(c\vert x)$ & $20.81 \pm 0.12$ & $7.49 \pm 0.09$ \\
          \bottomrule
    \end{tabular}
    \caption{Sampling with only one term from Equation \eqref{eq:latent_sampling} on CIFAR-10 dataset.}\label{tab:diff_term}    
\end{table}

\begin{figure}[h!]
    \centering
    \subfigure[Input]{\includegraphics[width=0.155\linewidth]{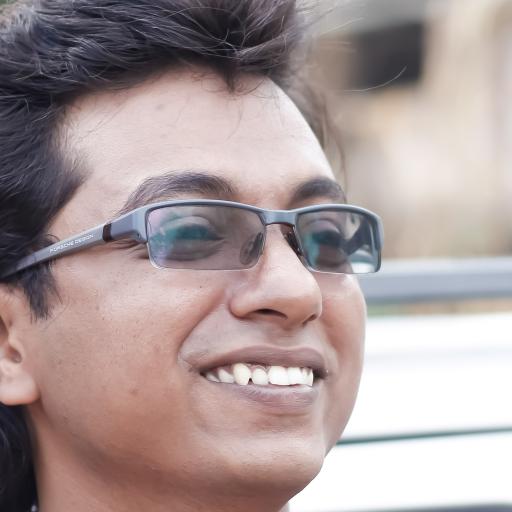}}
    \subfigure[Masked]{\includegraphics[width=0.155\linewidth]{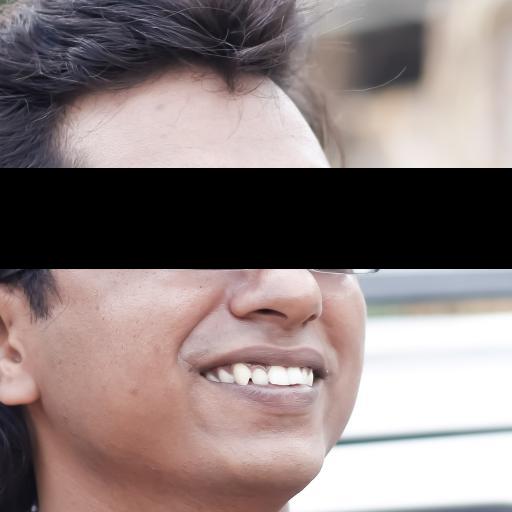}}
    \subfigure[Completed Images]{\includegraphics[width=0.64\linewidth]{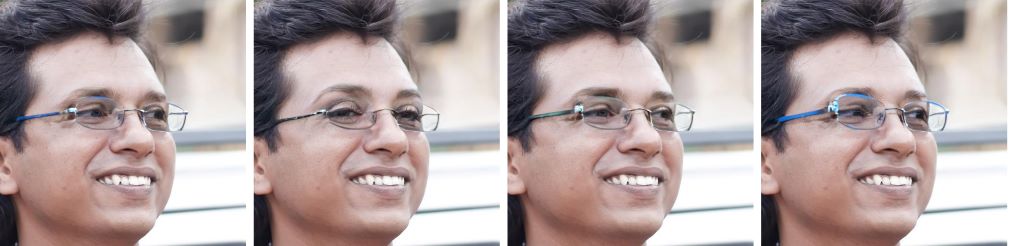}}
    \subfigure[Input]{\includegraphics[width=0.155\linewidth]{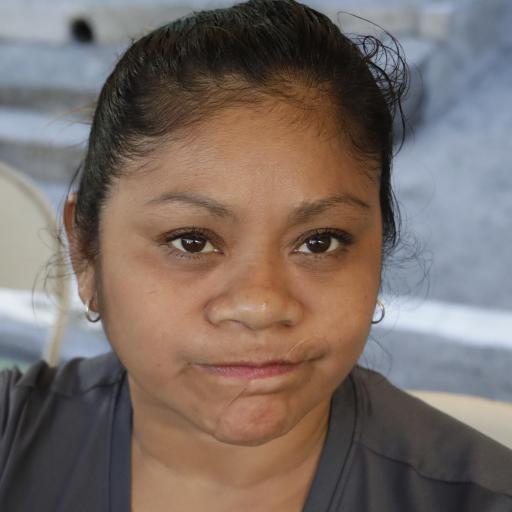}}
    \subfigure[Masked]{\includegraphics[width=0.155\linewidth]{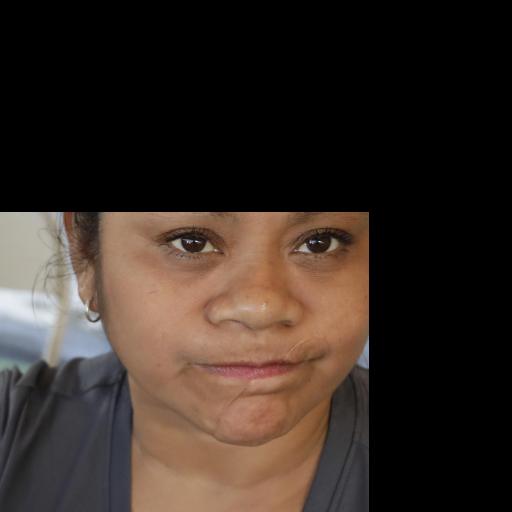}}
    \subfigure[Completed Images]{\includegraphics[width=0.64\linewidth]{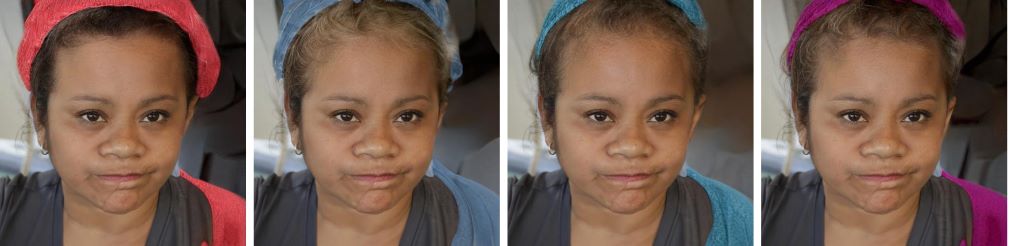}}
    \caption{Image inpainting examples on FFHQ ($512 \times 512$). The larger the mask, the more uncertainty thus more variations we observe.}
    \label{fig:Image_inpainting}
\end{figure}

\paragraph{Qualitative results on image inpainting} Some inpainting examples are provided in Figure \ref{fig:Image_inpainting}.

\begin{figure*}[h!]
    \centering
    \includegraphics[width=0.99\linewidth]{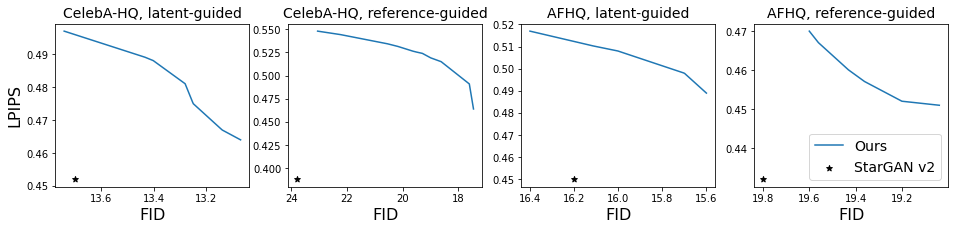}
    \caption{Detailed results in image translation task on Celeba-HQ ($256 \times 256$) and AFHQ ($256 \times 256$).}
    \label{fig:img_translation_fid_lpips}
\end{figure*}

\paragraph{Trade-off between FID and LPIPS in image translation} As mentioned earlier, we find these is a trade-off between FID and LPIPS in the image translation task. We present this their relationship in Figure \ref{fig:img_translation_fid_lpips}, where the plot are obtained by using different step size and update steps in our proposed method.

\begin{figure}[h!]
    \centering
    \subfigure[Input]{\includegraphics[width=0.105\linewidth]{figures/img_editing/ffhq_12.png}}
    \subfigure["face with straight hair", StyleCLIP-G]{\includegraphics[width=0.42\linewidth]{figures/img_editing/ffhq_12_straight_styleclip.png}}
    \subfigure["face with straight hair", ours]{\includegraphics[width=0.42\linewidth]{figures/img_editing/ffhq_12_straight_ours.png}}
    \subfigure[Input]{\includegraphics[width=0.105\linewidth]{figures/img_editing/ffhq_423.png}}
    \subfigure["face with beard", StyleCLIP-G]{\includegraphics[width=0.42\linewidth]{figures/img_editing/ffhq_423_beard_styleclip.png}}
    \subfigure["face with beard", ours]{\includegraphics[width=0.42\linewidth]{figures/img_editing/ffhq_423_beard_ours.png}}
        \subfigure[Input]{\includegraphics[width=0.105\linewidth]{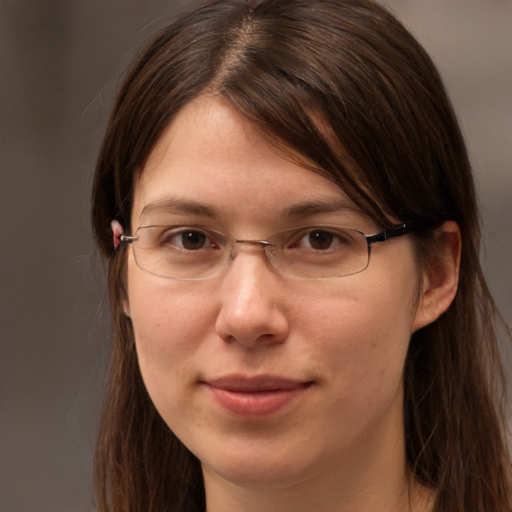}}
    \subfigure["face with curly hair", StyleCLIP-G]{\includegraphics[width=0.42\linewidth]{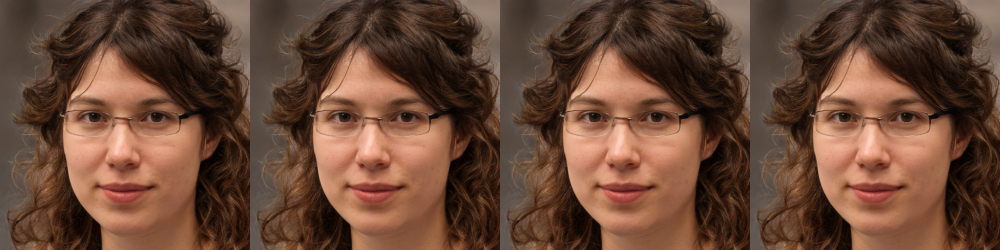}}
    \subfigure["face with curly hair", ours]{\includegraphics[width=0.42\linewidth]{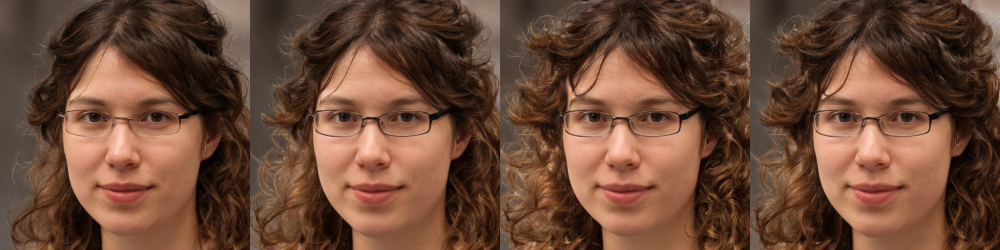}}
        \subfigure[Input]{\includegraphics[width=0.105\linewidth]{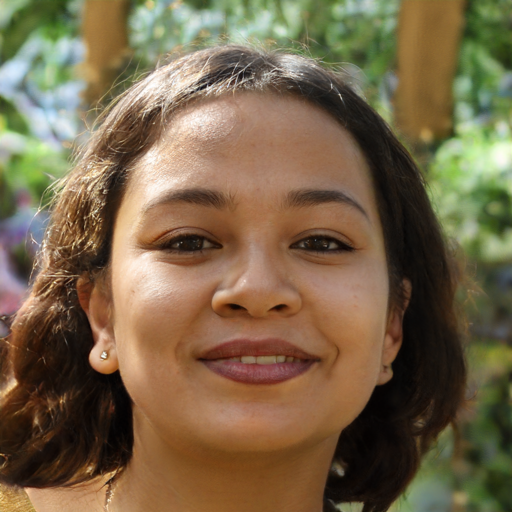}}
    \subfigure["face with short hair", StyleCLIP-G]{\includegraphics[width=0.42\linewidth]{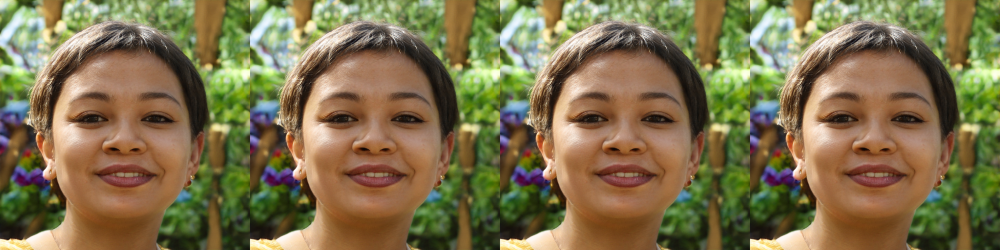}}
    \subfigure["face with short hair", ours]{\includegraphics[width=0.42\linewidth]{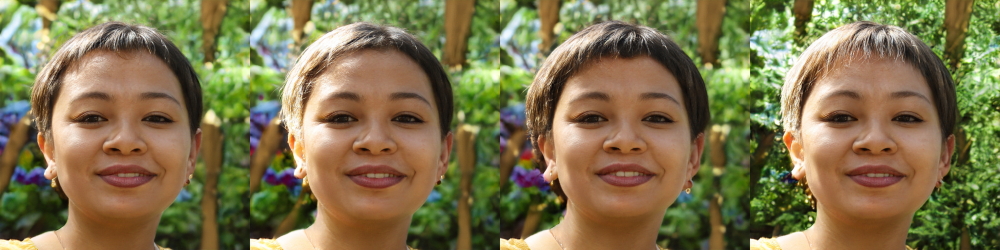}}    
    \caption{Text-guided image editing on FFHQ ($1024 \times 1024$) with diversity.} \label{fig:image_editing}
\end{figure}

\paragraph{Qualitative results on text-guided image editing} Some examples on text-guided image editing are provided in Figure \ref{fig:image_editing}, where we compare the proposed method with StyleCLIP-Global.

\end{document}